\documentclass{article}

\usepackage{macros}
\usepackage{xcolor}

\newcommand{\riccardo}[1]{{\color{orange}{[Riccardo] #1}}}

\usepackage{microtype}
\usepackage{graphicx}
\usepackage{subfigure}
\usepackage{booktabs, enumerate} 

\usepackage{hyperref}

\usepackage[accepted]{icml2018}

\usepackage{amsmath}
\usepackage{amsfonts}
\usepackage{breqn}
\usepackage{float}

\icmltitlerunning{Optimizing learning in variational autoencoders, with Kullback-Leibler and Renyi cdf bounds, for the continuous MNIST dataset}

\begin{document}

\twocolumn[
\icmltitle{Learning in Variational Autoencoders  \\with Kullback-Leibler and Renyi Integral Bounds}





\begin{icmlauthorlist}
\icmlauthor{Septimia S\^{a}rbu}{RIST}
\icmlauthor{Riccardo Volpi}{RIST}
\icmlauthor{Alexandra Pe\c{s}te}{RIST,MPL}
\icmlauthor{Luigi Malag\`{o}}{RIST}
\end{icmlauthorlist}

\icmlaffiliation{RIST}{Romanian Institute of Science and Technology, Cluj-Napoca, Romania}
\icmlaffiliation{MPL}{Max Planck Institute for Mathematics in the Sciences, Leipzig, Germany}
\icmlcorrespondingauthor{Septimia S\^{a}rbu}{sarbu@rist.ro}



\icmlkeywords{Machine Learning, ICML}

\vskip 0.3in
]


\printAffiliationsAndNotice{}  


\begin{abstract}
In this paper we propose two novel bounds for the log-likelihood based on Kullback-Leibler and the R\'{e}nyi divergences, which can be used for variational inference and in particular for the training of Variational AutoEncoders. Our proposal is motivated by the difficulties encountered in training VAEs on continuous datasets with high contrast images, such as those with handwritten digits and characters, where numerical issues often appear unless noise is added, either to the dataset during training or to the generative model given by the decoder. The new bounds we propose, which are obtained from the maximization of the likelihood of an interval for the observations, allow  numerically stable training procedures without the necessity of adding any extra source of noise to the data.
\end{abstract}

\section{Introduction}
\label{intro}
Variational AutoEncoders (VAEs) \cite{Kingma2014}, \cite{Rezende2014} are a class of generative models, providing a powerful approach to conduct statistical inference with complex probabilistic models, within the variational Bayesian framework. As any autoencoder, VAE is composed of an encoder and a decoder. The encoder maps the high-dimensional input data into a lower-dimensional latent model, through a probabilistic model $q(z|x)$. The decoder does the reverse, by modeling a conditional probability distribution $p(x|z)$ that generates the reconstructed input, as the output. The encoder and the decoder are trained in synergy to provide reconstructed images of high quality, constrained to the assumption that the latent variable $z$ is distributed as a chosen prior $p(z)$ on the latent space. After training, VAE can act as a generative model, by sampling from the prior distribution in the latent space and, then, decoding these samples. Both the decoder and the encoder conditional probabilities are parameterized by a neural network and are generally optimized through maximizing a lower bound on the model evidence, known in the literature as the ELBO \cite{Jordan1999}.

Recently, three main research directions have been pursued, to provide better learning and, thus, better performance in VAE: increasing the complexity of the approximate posterior, developing more accurate lower bounds to the maximum likelihood as training objectives and adding noise at different stages in the algorithmic framework. The performance of VAE can be improved through the use of more complex models for the approximate posterior:  normalizing flows \cite{Rezende2015}, \cite{Kingma2016} and products of conditional distributions \cite{Sonderby2016}. The latter is employed both in the model of the approximate posterior distribution, $q(z|x)$, as well as in the one of the generative distribution, $p(x|z)$.

In the context of developing bounds on the model evidence, the original ELBO bound, as the objective function to train VAE, is derived using the Kullback-Leibler divergence between the true and approximate posteriors \cite{Kingma2014}. This bound is tightened in \cite{Burda2016}, through an importance weighted unbiased estimator of the marginal distribution, $p(x)$, and its gradient estimator is given in \cite{Mnih2016}. The state-of-the-art results provided by IWAE are enhanced through several linear combinations of VAE and IWAE \cite{Rainforth2018}. In addition, the authors of \cite{Rainforth2018} investigate under which conditions tighter bounds improve the learning and generative capabilities of VAE.

Another extension of the ELBO is derived starting from the R\'{e}nyi $\alpha$-divergence between the true and approximate posteriors \cite{Li2016}. Its importance weighted estimator is given in \cite{Webb2016}. With the $\chi^2$ divergence, an upper bound, termed CUBO, is proposed in \cite{dieng2017variational} and the gap between this newly introduced upper bound, and the original ELBO becomes the training objective.

In addition to developing more accurate bounds, learning in VAEs and, in general, in autoencoders is also facilitated by adding noise. Stochastic noise corruption, applied to the input data, is able to train denoising autoencoders and to generate samples of high visual quality \cite{Vincent2010}, \cite{Bengio2013}. Bit-flip and drop-out noise added to the input layer of the encoder of a VAE and Gaussian noise added to the samples given by the encoder is fundamental to perform inference in the case of new data inputs \cite{Rezende2014}. By adding noise to the latent model and by proposing a new tractable bound, called the denoising variational lower bound, the authors of \cite{Im2017} obtain improved performance over VAE and IWAE. In particular, they transform the approximate posterior using a noise distribution and thus obtain a more complex model of the approximate posterior based on a Gaussian mixture model. With a plethora of possible approaches to chose from, in order to successfully train VAE on continuous datasets and obtain good reconstructions, we feel the need of a deeper understanding and a thorough study of the role of the bounds on the model evidence and the added noise at different steps in the training procedure. This work-in-progress paper will present our first steps in this direction and it will focus on the analysis of training VAE over the continuous MNIST and OMNIGLOT datasets.

\subsection{Problem Statement}
\label{sec:elbounbalance}
\begin{figure}
\begin{center}
\centerline{\includegraphics[width=0.8\columnwidth]{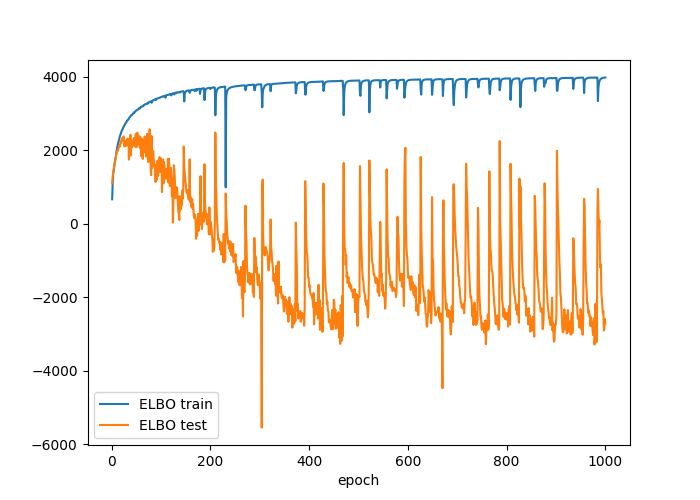}}
\caption{The ELBO as a function of the number of epochs, for train and test continuous MNIST images.}
\label{fig_MNIST_elbo_bound}
\end{center}
\end{figure}
\begin{figure}
\begin{center}
\centerline{\includegraphics[width=0.8\columnwidth]{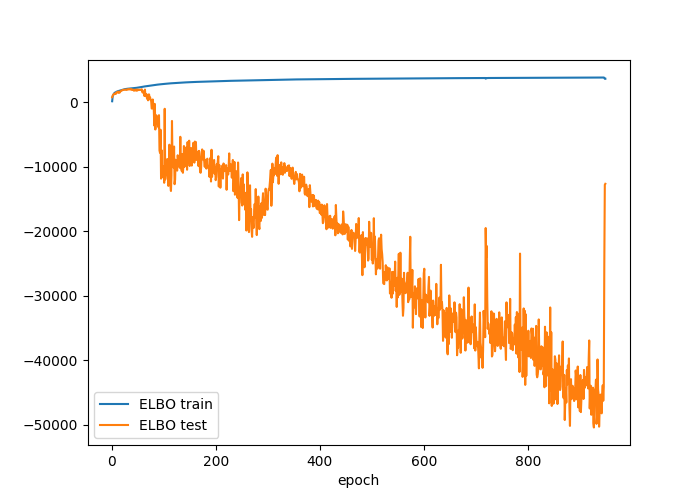}}
\caption{The ELBO as a function of the number of epochs, for train and test continuous OMNIGLOT images.}
\label{fig_OMNIGLOT_elbo_bound}
\end{center}
\end{figure}
\begin{figure}
\vskip 0.2in
\begin{center}
\centerline{\includegraphics[width=\columnwidth]{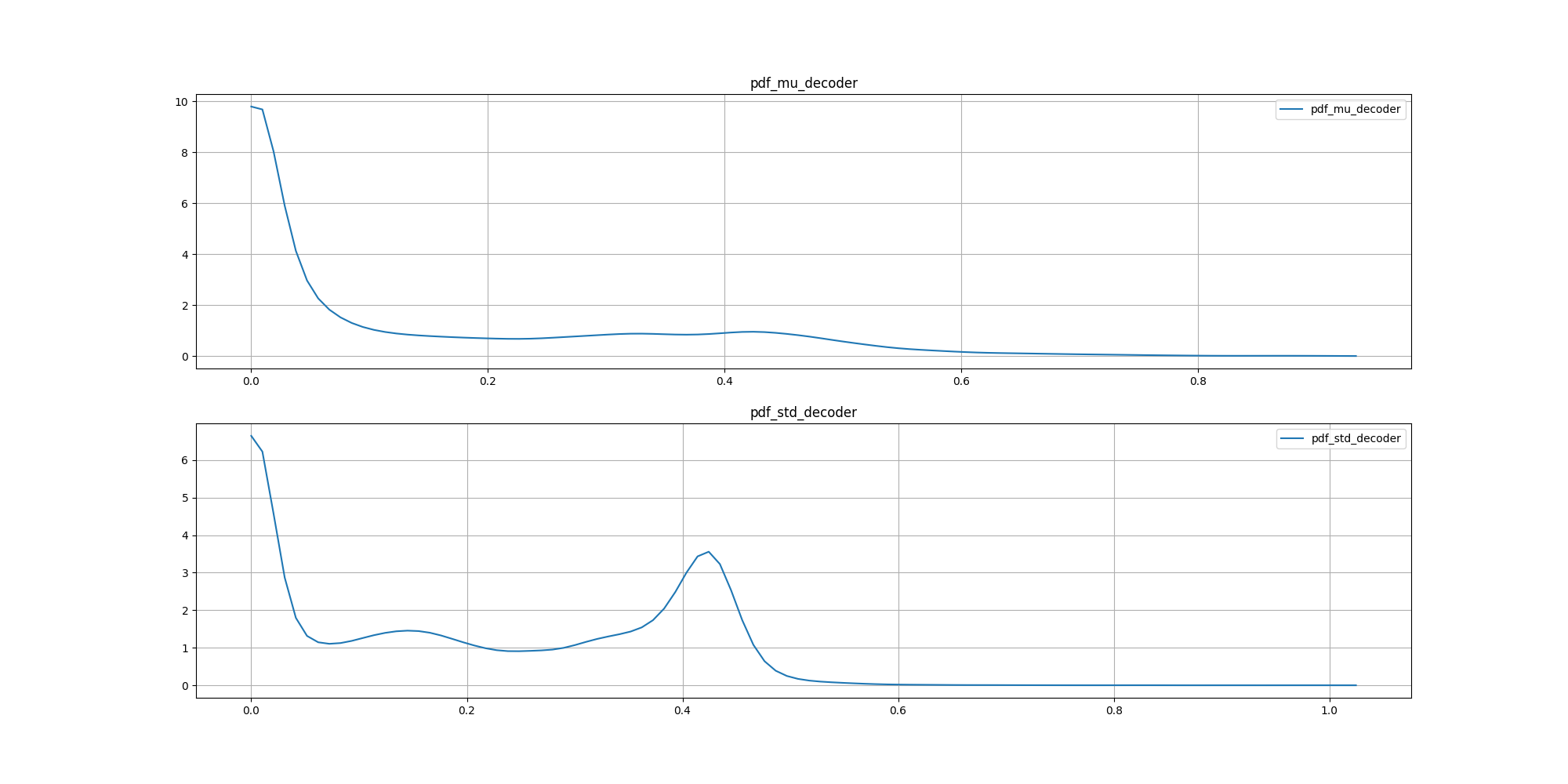}}
\caption{The distribution of the mean and the standard deviation of the decoder, after $30$ epochs and $550$ steps, during training.}
\label{prob_stat}
\end{center}
\vskip -0.2in 
\end{figure}

This paper stems from the difficulties encountered when training a VAE for continuous images with high contrast, such as the handwritten digits and characters from the MNIST and OMNIGLOT datasets, where the majority of the pixels are concentrated close to $0$ or $1$. In the last layer of the decoder, we used the sigmoid activation function for the mean and the exponential activation function for the standard deviation, to ensure its positivity, and a Gaussian independence model for the generative distribution. The standard ELBO was the objective function used for training.\\
During the learning process, we observed that many of the standard deviations given by the decoder become very close to $0$, as shown in Figure \ref{prob_stat}. This phenomenon results in the densities, $p(x|z)$ growing to very large values, which implies that the reconstruction error of the ELBO bound gets much bigger than the KL term with the prior. This creates an unbalance in the two terms composing the bound and destabilize the optimization procedure, figs.~\ref{fig_MNIST_elbo_bound}-\ref{fig_OMNIGLOT_elbo_bound} . This means that the reward by minimizing the standard deviations is much bigger than the reward of increasing the Kullback-Leibler term of the bound, which could potentially impact the learning of the prior distribution. As a solution to this problem, in the maximum likelihood framework, we propose to maximize the likelihood of intervals for the observations, instead of standard likelihood. 

Our contribution is the derivation of novel integral lower bounds on the model evidence, by taking into account that in any stochastic model, probabilities are bounded by $1$, whereas densities can tend to infinity. In order to mitigate the above mentioned issues, we construct a learning objective based on maximizing the likelihood of an interval for the observations, instead of maximizing the standard likelihood. The new learning objective represents a lower bound on the model evidence. We derive this lower bound in two forms: one starting from the Kullback-Leibler divergence between the true and approximate posterior distributions and another one from the R\'{e}nyi $\alpha$-divergence between the same distributions. We provide proof-of-concept results of the training algorithms constructed with these novel bounds. 

We developed our idea of using likelihoods of intervals independently from \cite{Salimans2017}. The authors of \cite{Salimans2017} improve the conditional densities model of \cite{vanOord2016b}, by introducing a latent model for the color intensity of the pixels, which is given by a mixture of logistic distributions. The logarithm of the probability of an interval around the currently observed pixel, in discretized form, becomes the objective function. With this improved architecture, they obtain state-of-the-art result on the CIFAR-$10$ benchmark dataset. Their purpose was to alleviate the memory and computational burden of the PixelCNN algorithm, as well as to obtain better results and a speed-up in training. In contrast, our motivation was to solve an unboundedness problem of the probability densities, which appears during the maximization of the objective function.

\section{Mathematical Derivations}
\label{math}
In this section, we provide the technical derivations for the newly introduced Kullback-Leibler and R\'{e}nyi integral bounds, termed IELBO and IRELBO, respectively.
\subsection{The Integral ELBO}
We start the derivation of the IELBO bound from the definition of the model evidence:
\begin{align}
\int_{a}^{b} p(x) \mathrm{dx} &=\int_{a}^{b} \left(\int_{z} p(x|z) \cdot p(z) \mathrm{dz} \right) \mathrm{dx} \nonumber \\
&=\int_{a}^{b} \left(\int_{z} \frac{p(x|z) \cdot p(z)}{q(z|x_i)} \cdot q(z|x_i) \mathrm{dz} \right) \mathrm{dx} \nonumber \\
&=\int_{z} \frac{\left( \int_{a}^{b} p(x|z) \mathrm{dx} \right) \cdot p(z)}{q(z|x_i)} \cdot q(z|x_i)   \mathrm{dz} \nonumber \\
&=\mathbb{E}_{q(z|x_i)} \left[ \frac{\left( \int_{a}^{b} p(x|z) \mathrm{dx} \right) \cdot p(z)}{q(z|x_i)} \right]. \nonumber
\end{align}
We would like to point out that the approximate posterior, $q(z|x)$, plays the role of an importance distribution and can be chosen to be any probability density function. Here, in particular, we choose to fix the conditional variable to an example $x_i$, $q(z|x_i)$.

Applying Jensen's inequality and taking $b=x_i+\epsilon$ and $a=x_i-\epsilon$ , we obtain the following lower bound:
\begin{align}
\log{\int_{x_i-\epsilon}^{x_i+\epsilon} p(x)\mathrm{dx}} &\geq \mathbb{E}_{q(z|x_i)} \left[\log{ \frac{\left( \int_{x_i-\epsilon}^{x_i+\epsilon} p(x|z) \mathrm{dx} \right) \cdot p(z)}{q(z|x_i)}} \right] \nonumber \\
&=\mathbb{E}_{q(z|x_i)} \left[\log{\left( \int_{x_i-\epsilon}^{x_i+\epsilon} p(x|z) \mathrm{dx} \right)}\right]- \nonumber \\
&- D_{KL}[q(z|x_i)||p(z)].
\end{align}

We define the integral ELBO bound as
\begin{align}
\text{IELBO} &= \mathbb{E}_{q(z|x_i)} \left[\log{\left( \int_{x_i-\epsilon}^{x_i+\epsilon} p(x|z) \mathrm{dx} \right)}\right]- \nonumber \\
& - D_{KL}[q(z|x_i)||p(z)].
\end{align}

\subsection{The Integral R\'{e}nyi ELBO}
We start the derivation of the IRELBO bound from R\'{e}nyi's $\alpha$-divergence \cite{Renyi1961} between the approximate posterior distribution, $q(z|x_i)$, conditioned on the current example, $x_i$, and the true posterior distribution, $p(z|x)$:
\begin{align}
D_{\alpha}[q(z|x_i)||p(z|x)]&=\frac{1}{\alpha-1} \log{\int_{z} q(z|x_i)^{\alpha} \cdot p(z|x)^{1-\alpha}} \mathrm{dz}. \nonumber
\end{align}
The definite integral of this divergence in an arbitrary interval, $a$ and $b$, reads:
\begin{align}
&\int_{a}^{b} (\alpha-1) \cdot D_{\alpha}[q(z|x_i)||p(z|x)] \enskip \mathrm{dx} = \nonumber \\
&=\int_{a}^{b} \log \left[ \int_{z} q(z|x_i)^{\alpha} \cdot p(z|x)^{1-\alpha} \mathrm{dz} \right] \mathrm{dx} \nonumber \\
&\leq (b-a) \cdot \log{\left( \frac{1}{b-a} \cdot C \right)},
\end{align}
where the last line follows from Jensen's inequality and
\begin{align}
&C = \int_{a}^{b} \int_{z} q(z|x_i)^{\alpha} \cdot p(z|x)^{1-\alpha} \mathrm{dz} \enskip \mathrm{dx}. \nonumber \\
&= \int_{a}^{b} \left[ \int_{z} \left( \frac{q(z|x_i) \cdot p(x)}{p(x|z) \cdot p(z)} \right)^{\alpha-1} \cdot q(z|x_i) \mathrm{dz} \right] \mathrm{dx} \nonumber \\
&= \mathbb{E}_{q(z|x_i)} \left[\int_{a}^{b}  p(x)^{\alpha-1} \cdot \left( \frac{q(z|x_i) }{p(x|z) \cdot p(z)} \right)^{\alpha-1} \mathrm{dx} \right]. \nonumber
\end{align}
\begin{align}
\text{Let } f(x)&=p(x)^{\alpha-1} \geq 0 \nonumber \\
g(x)&= \left( \frac{q(z|x_i) }{p(x|z) \cdot p(z)} \right)^{\alpha-1} \geq 0 . \nonumber \\
\Rightarrow C &= \mathbb{E}_{q(z|x_i)} \left[ \int_{a}^{b} f(x) \cdot g(x) \mathrm{dx} \right].
\end{align}

Applying H\"{o}lder's inequality in integral form yields
\begin{align}
&\int_{a}^{b} f(x) \cdot g(x) \mathrm{dx}=\int_{a}^{b} |f(x) \cdot g(x)| \mathrm{dx} \nonumber \\
&\leq \left(\int_{a}^{b} |f(x)|^u \mathrm{dx} \right)^{\frac{1}{u}} \cdot \left(\int_{a}^{b} |g(x)|^w \mathrm{dx} \right)^{\frac{1}{w}} \nonumber \\
&= \left(\int_{a}^{b} f(x)^u \mathrm{dx} \right)^{\frac{1}{u}} \cdot \left(\int_{a}^{b} g(x)^w \mathrm{dx} \right)^{\frac{1}{w}}, \nonumber \\
& \text{ with } u,w > 1, \frac{1}{u}+\frac{1}{w}=1.
\end{align}

For $0 \leq \alpha-1<1$, if we choose $u=\frac{1}{\alpha-1}>1$, then $w=\frac{1}{2-\alpha}>1$ and
\begin{align}
C &\leq \mathbb{E}_{q(z|x_i)} \left[ \left(\int_{a}^{b} p(x) \mathrm{dx} \right)^{\alpha-1} \cdot \right. \nonumber \\
& \left. \cdot \left(\int_{a}^{b} \left( \frac{q(z|x_i) }{p(x|z) \cdot p(z)} \right)^{\frac{\alpha-1}{2-\alpha}} \mathrm{dx} \right)^{2-\alpha} \right] \nonumber \\
& =  \left(\int_{a}^{b} p(x) \mathrm{dx} \right)^{\alpha-1} \cdot \nonumber \\
& \cdot \mathbb{E}_{q(z|x_i)} \left\{\left[\left( \frac{q(z|x_i) }{p(z)} \right)^{\frac{\alpha-1}{2-\alpha}} \cdot \int_{a}^{b} \left( \frac{1}{p(x|z)} \right)^{\frac{\alpha-1}{2-\alpha}} \mathrm{dx} \right]^{2-\alpha}  \right\} \nonumber \\
& =  \left(\int_{a}^{b} p(x) \mathrm{dx} \right)^{\alpha-1} \cdot \nonumber \\
& \cdot \mathbb{E}_{q(z|x_i)} \left\{ \left[ \int_{a}^{b}  p(x|z)^{\frac{1-\alpha}{2-\alpha}} \mathrm{dx} \right]^{2-\alpha} \cdot \left[ \frac{q(z|x_i) }{p(z)} \right]^{\alpha-1} \right\}. \nonumber
\end{align}
\begin{align}
&\text{Let } h(z,x_i)=\left[ \int_{a}^{b} p(x|z)^{\frac{1-\alpha}{2-\alpha}} \mathrm{dx} \right]^{2-\alpha} \cdot \left[\frac{q(z|x_i)}{p(z)} \right]^{\alpha-1}. \nonumber \\
& \Rightarrow \log{C} \leq  (\alpha-1) \cdot \log{\left(\int_{a}^{b} p(x) \mathrm{dx} \right)} + \nonumber \\
& +\log{\mathbb{E}_{q(z|x_i)} \left[h(z,x_i) \right] }. \nonumber
\end{align}
For $1 < \alpha < 2$, we have that:
\begin{align}
0 &\leq \int_{a}^{b} (\alpha-1) \cdot D_{\alpha}[q(z|x_i)||p(z|x)] \enskip \mathrm{dx} \nonumber \\
&\leq (b-a) \cdot \left[\log{\frac{1}{b-a}} + (\alpha-1) \log{\int_{a}^{b} p(x)\mathrm{dx}} + \right. \nonumber \\
& \left. +  \log{\mathbb{E}_{q(z|x_i)} \left[ h(z,x_i) \right]} \right],  \nonumber
\end{align}
\begin{align}
\Rightarrow \log{\int_{a}^{b} p(x)\mathrm{dx}} \geq \frac{1}{\alpha-1} \cdot \log{\frac{b-a}{\mathbb{E}_{q(z|x_i)} \left[ h(z,x_i) \right]}} ,  \nonumber \\
\Rightarrow \log{\int_{x_i-\epsilon}^{x_i+\epsilon} p(x)\mathrm{dx}} \geq \frac{1}{\alpha-1} \cdot \log{\frac{2 \cdot \epsilon}{\mathbb{E}_{q(z|x_i)} \left[ h(z,x_i) \right]}}. \nonumber
\end{align}
We define the integral R\'{e}nyi bound as
\begin{align}
\text{IRELBO} = \frac{1}{\alpha-1} \cdot \log{\frac{b-a}{\mathbb{E}_{q(z|x_i)} \left[ h(z,x_i) \right]}}.
\end{align}
In our assumption, $p(x|z)$ is an independence model, with each pixel, $x^{(j)}$ distributed as $\mathcal{N}(\mu_d^{(j)},\sigma_d^{2(j)})$. Because this distribution is raised to a negative exponent, we require the Dawson function, $D(x)=e^{-x^2} \cdot \int_0^x e^{t^2} \mathrm{dt}$, to compute the integral
\begin{align}
\int_{a}^{b} p(x|z)^{\frac{1-\alpha}{2-\alpha}} \mathrm{dx}&=\prod_j\left(\sqrt{2\pi \sigma_d^{2(j)} } \right)^{\frac{1-\alpha}{2-\alpha}} \cdot \sqrt{2 \cdot \frac{2-\alpha}{\alpha-1} \cdot \sigma_d^{2(j)}} \cdot \nonumber \\
& \cdot \left[ D(b_t) \cdot e^{b_t^2}-D(a_t) \cdot e^{a_t^2} \right],
\end{align}
with $b_t=\left(b-\mu_d^{(j)} \right)/\left(\sqrt{2 \cdot \frac{2-\alpha}{\alpha-1} \cdot \sigma_d^{2(j)} } \right)$ and $a_t=\left(a-\mu_d^{(j)} \right)/\left(\sqrt{2 \cdot \frac{2-\alpha}{\alpha-1} \cdot \sigma_d^{2(j)} } \right)$.
\section{Experimental Setting}
\label{experim}
\subsection{The Model}
An image is represented by pixel values in a certain range. In maximum log likelihood estimation, we replace the $\log{p(x)}$ with the likelihood of the interval $\log{\int_{a}^{b}} p(x) \mathrm{dx}$, where $a,b$ can be taken such that $b=x_i+\epsilon$ and $a=x_i-\epsilon$, with $\epsilon\in\reals^n$ and small enough, such that both $a,b\in [0,1]^n$. The value of $\epsilon$ can be seen as expressing a numerical tolerance on the value of the pixels and in practice it is motivated by the fact that pixels printed on screen have limited precision. 

In VAEs we assume that it is possible to find a latent variable $z\in \reals^k$ that could in principle explain the observation $x\in \reals^n$. To reduce the comparison to the essential, we will analyze the simplest model present in the literature, composed of a Gaussian approximate posterior with a diagonal covariance matrix and a standard Gaussian distribution for the prior in the latent space. 
For the encoding process, the input $x$ is passed to a neural network with output the mean and the logarithm of the covariances of a Gaussian distribution in the latent space, i.e. $q(z|x) = \mathcal{N}(\mu_x, \Sigma_x)$. The positivity of the entries of the covariance matrix is guaranteed by the exponential transformation on the output of the network. For the decoder, we will use an analogous model: the output of the network neural network given by the mean and covariance of a Gaussian probability distribution for the observations, i.e., $p(x|z) = \mathcal{N}(\mu_z, \Sigma_z)$, where the sigmoid function $\sigma$ is used to restrict the mean between $0$ and $1$.

\subsection{Neural Network Architectures}
In the learning schemes for the Kullback-Leibler integral bound (IELBO) we use the following architecture for a VAE: the neural network of the encoder and that of the decoder contain two deterministic hidden layers, each with $200$ nodes and the ELU activation function. The standard deviation given by the encoder is transformed through the exponential function, to ensure its positivity. The dimension of the latent space is equal to $25$. The learning rate for the Adam optimizer is equal to $0.0001$. The integration interval is equal to $0.02$. All images have been rescale to $[0.001,0.999]$ to avoid numerical issues. We consider $10$ samples drawn from the approximate posterior $q(z|x)$ and a batch size of $100$ input samples. The weights are initialized with the xavier initialization, while the biases are set equal to $0.1$.

In the learning schemes for the R\'{e}nyi integral bound, IRELBO, we use the following architecture of VAE: the neural network of the encoder and that of the decoder contain one hidden layer, with $400$ nodes and the ReLU activation function. The standard deviation given by the encoder is transformed through the exponential function, to ensure its positivity. The dimension of the latent space is equal to $20$ and the learning rate for Adam is equal to $0.0005$. The integration interval is equal to $1$. We consider $\alpha=1.5$, one sample drawn from the approximate posterior $q(z|x)$ and a batch size of $100$ input samples. The weights are initialized with the xavier initialization. 

\section{Experimental Results and Discussion}
\label{diss}
In this section we provide preliminary results for the newly introduced integral bounds employed for the training of a standard VAE, for the continuous MNIST \cite{LeCun1998} and OMNIGLOT datasets \cite{Lake2015}. No extra source of noise has been added during training, neither to the input samples, nor to generative Gaussian distribution $p(x|z)$, for instance in the form of a lower bound for the entries of the covariance matrix. The Figures in the paper which shows reconstructed test images represents in the first and fourth rows the original images; in the second and fifth rows the reconstructed images using the mean of the decoder distribution (obtained by encoding the mean of $q(z|x)$); finally in the third and sixth rows the reconstructed images using the mean of the decoder distribution (obtained by passing a sample drawn from $q(z|x)$ through the decoder).
\begin{figure}[ht]
\vskip 0.2in
\begin{center}
\centerline{\includegraphics[width=\columnwidth]{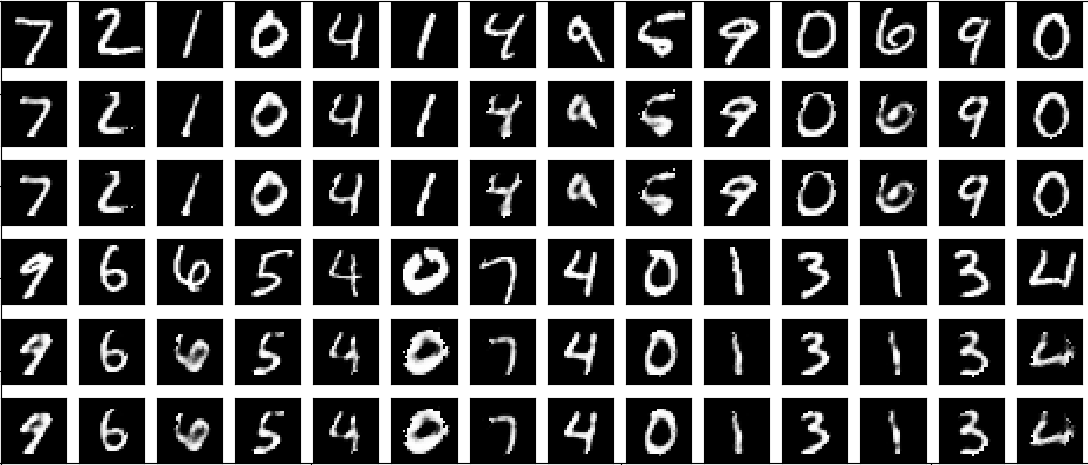}}
\caption{The test continuous MNIST images reconstructed with the original ELBO, after $1000$ epochs.}
\label{fig_MNIST_elbo_images}
\end{center}
\vskip -0.2in
\end{figure}

\begin{figure}[ht]
\vskip 0.2in
\begin{center}
\centerline{\includegraphics[width=0.8\columnwidth]{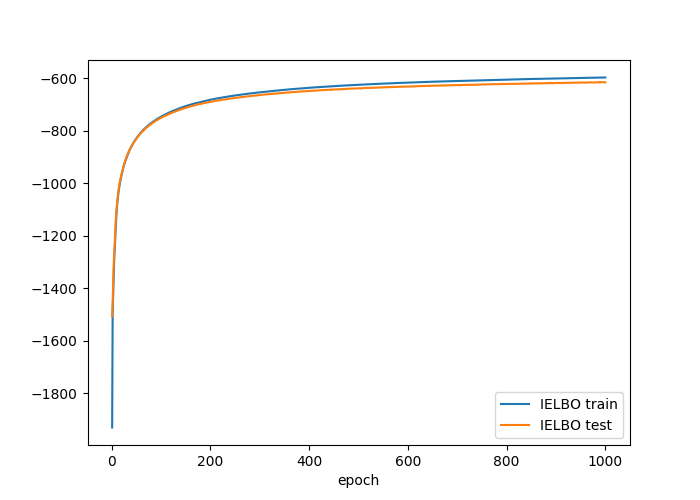}}
\caption{The IELBO as a function of the number of epochs, for train and test continuous MNIST images.}
\label{fig_MNIST_ielbo_bound}
\end{center}
\vskip -0.2in
\end{figure}

\begin{figure}[ht]
\vskip 0.2in
\begin{center}
\centerline{\includegraphics[width=\columnwidth]{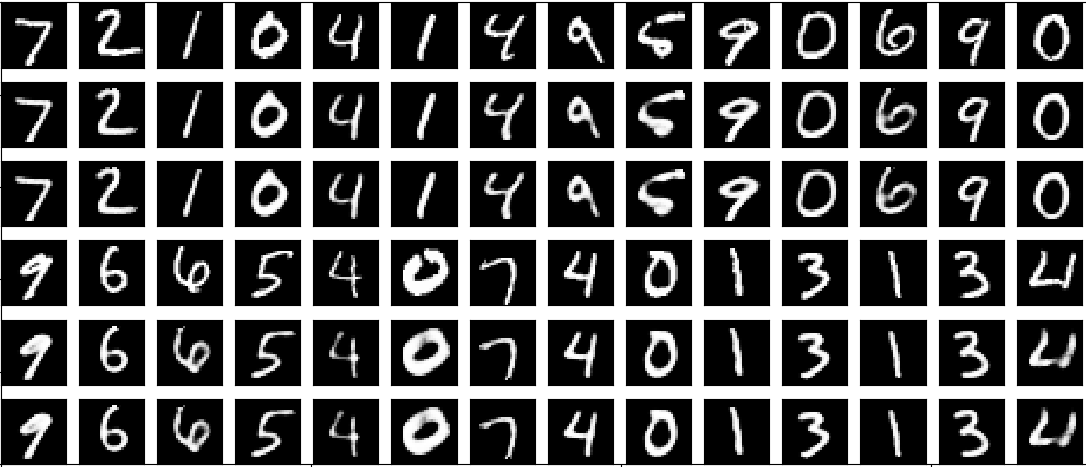}}
\caption{The test continuous MNIST images reconstructed with the IELBO, after $1000$ epochs.}
\label{fig_MNIST_ielbo_images}
\end{center}
\vskip -0.2in
\end{figure}

\begin{figure}[ht]
\vskip 0.2in
\begin{center}
\centerline{\includegraphics[width=\columnwidth]{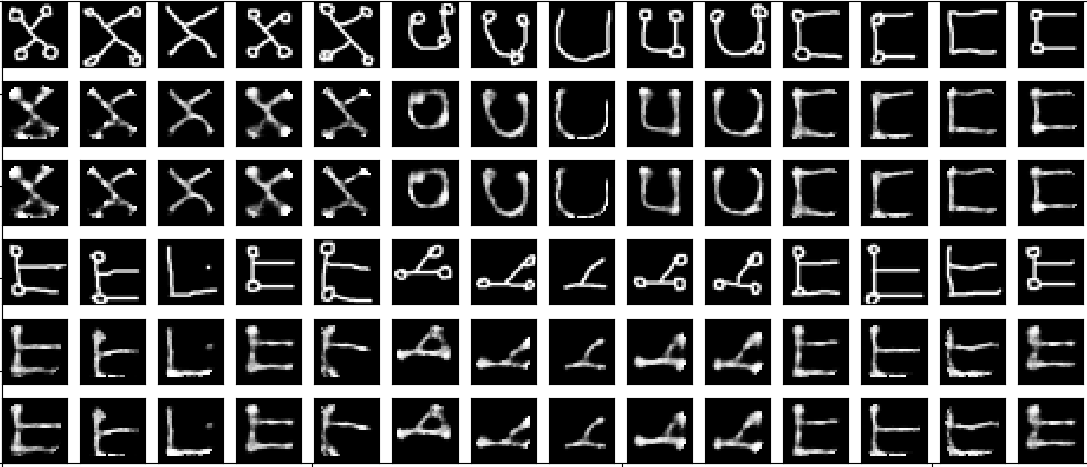}}
\caption{The test continuous OMNIGLOT images, reconstructed with the IELBO, after $1000$ epochs.}
\label{fig_OMNIGLOT_ielbo_images}
\end{center}
\vskip -0.2in
\end{figure}

\begin{figure}[ht]
\vskip 0.2in
\begin{center}
\centerline{\includegraphics[width=0.8\columnwidth]{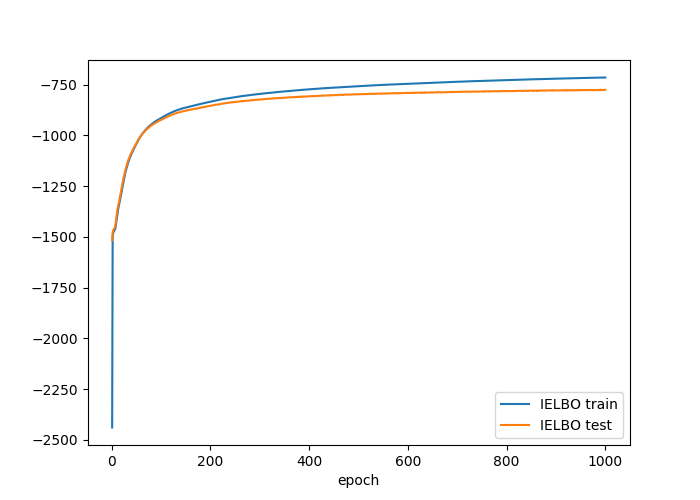}}
\caption{The IELBO as a function of the number of epochs, for train and test continuous OMNIGLOT images.}
\label{fig_OMNIGLOT_ielbo_bound}
\end{center}
\vskip -0.2in
\end{figure}

The results we have presented indicate that the original ELBO is not suited for high contrast images, unless noise is added either to the data or to the model. Using the IELBO, we were able to efficiently reconstruct images, as shown in Figures \ref{fig_MNIST_ielbo_bound}, \ref{fig_MNIST_ielbo_images}, \ref{fig_OMNIGLOT_ielbo_images}, and \ref{fig_OMNIGLOT_ielbo_bound}. The bounds on the train and test datasets saturate after few hundred epochs, are smooth and have comparable values. The quality of the reconstructed MNIST test images appear to be very good, while for the reconstructed OMNIGLOT test images is of acceptable quality. On these datasets, the original ELBO without added any extra source of noise performs poorly. For MNIST, Figure \ref{fig_MNIST_elbo_images} illustrates reconstructed test images of medium quality, while in Figure \ref{fig_MNIST_elbo_bound} it is possible to see that the bound evaluated on the test images has severe fluctuations associated to numerical issues, and deviates significantly from the one evaluated on the training set, as the epochs increase. For OMNIGLOT, we were not able to train the algorithm with the original ELBO, due to numerical issues as those in Figure \ref{fig_OMNIGLOT_elbo_bound}.

\begin{figure}[!ht]
\vskip 0.2in
\begin{center}
\centerline{\includegraphics[width=\columnwidth]{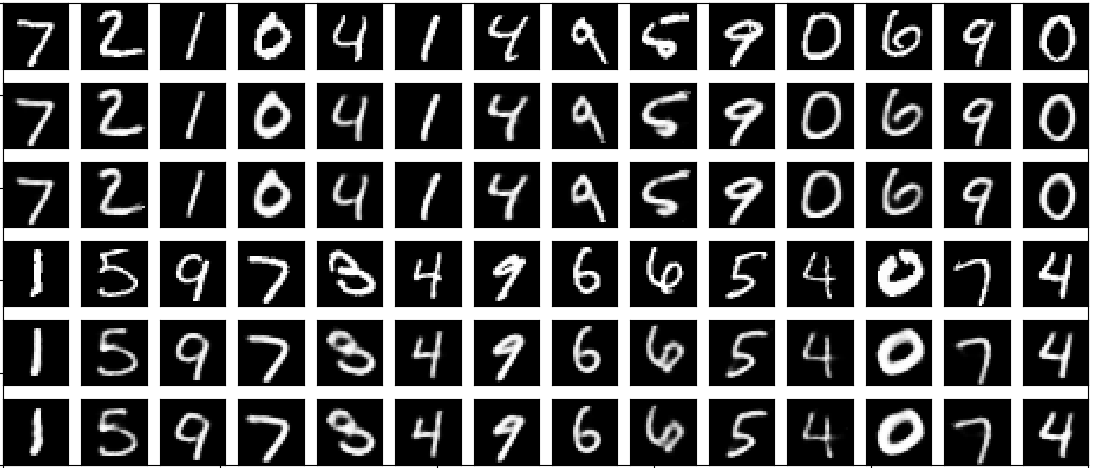}}
\caption{The test continuous MNIST images, reconstructed with the IRELBO, after $1000$ epochs.}
\label{fig_MNIST_irelbo_images}
\end{center}
\vskip -0.2in
\end{figure}

\begin{figure}[!ht]
\vskip 0.2in
\begin{center}
\centerline{\includegraphics[width=0.8\columnwidth]{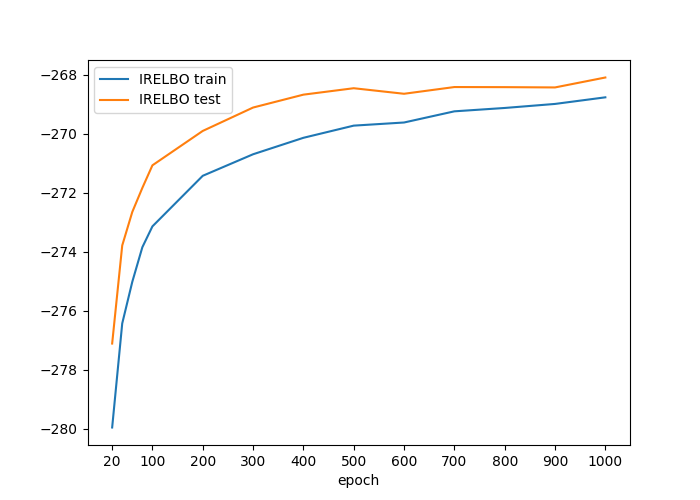}}
\caption{The IRELBO as a function of the number of epochs, for train and test continuous MNIST images.}
\label{fig_MNIST_irelbo_bound}
\end{center}
\vskip -0.2in
\end{figure}

Using the R\'{e}nyi integral bound, we obtained very good quality for the reconstructed MNIST test images in Figure \ref{fig_MNIST_irelbo_images}, and smooth bounds on the train and test examples, in Figure \ref{fig_MNIST_irelbo_bound}. 
Compared with the IELBO results on MNIST, training VAE with the IRELBO requires a greater number of epochs for convergence. 

The integral bounds contain an extra hyper-parameter compared to the original ELBO. In our experiments, we observed that the value of the bound and the reconstruction quality of the test images are significantly affected by the choice of the integration interval. 

As future work, we plan to study the impact of the size of the integration interval on the value of the bound and the quality of the reconstructed images. We plan to investigate how to efficiently select the integration interval for $\log{\int_a^b p(x)}$, for instance by dividing the $[0,1]$ interval in $k$ fixed bins, or by doing this dynamically based on the pixel mean.   
\begin{figure}[!ht]
\vskip 0.2in
\begin{center}
\centerline{\includegraphics[width=\columnwidth]{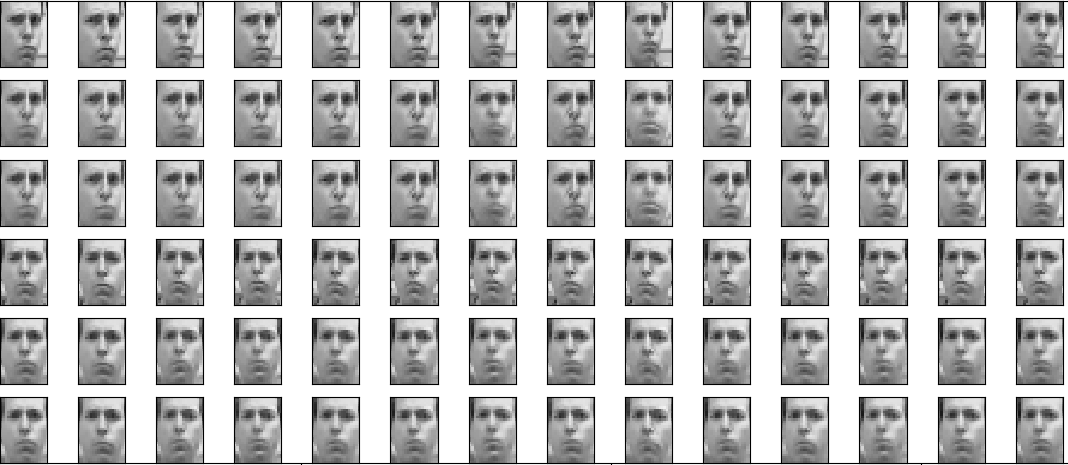}}
\caption{The test continuous Frey Faces images, reconstructed with the IELBO, after $1000$ epochs.}
\label{fig_FreyFaces_ielbo_images}
\end{center}
\vskip -0.2in
\end{figure}
Several relevant issues remain to be solved. We plan to study the behavior of these novel bounds on other continuous datasets, which have a more uniform distribution of the values of the pixels between $[0,1]$, in opposition to the continuous MNIST and OMNIGLOT datasets. Preliminary experiments on datasets with low contrast images, such as Frey Faces \cite{FreyFacelink}, provided good results. We are also investigating the impact on the reconstruction quality and on the value of the bound of different models for the decoder, such as the logit-normal distribution. 
Another relevant research direction is the study of the gradients of the novel bounds, from the point of view of the efficiency of the estimators and the speed of convergence. 
Finally, it will be also interesting to examine the advantages of the R\'{e}nyi integral bound over the Kullback-Leibler one, to determine if a more complicated divergence function provides benefits in the learning capacity of the algorithm.

\section{Conclusions}
\label{Conc}
Motivated by the numerical difficulties encountered during the training of VAEs using the standard ELBO for continuous datasets characterized by high contrast (non-binary) images, such as with MNIST and OMNIGLOT, we introduced two novel lower bounds for the log likelihood, computed by maximizing the likelihood of intervals for the continuous observations. 
We conducted proof-of-concept experiments, which showed the capacity of our algorithms to produce good quality reconstructed test images and avoid numerical issues, without the need to add extra noise either to the data during training, or to the generative model. 
One benefit of our bounds is that they require the computation of likelihoods of intervals, which implicitly prevent the generation of small variances for the reconstructed inputs. Indeed, the likelihood of an interval can be bounded, thus, avoiding the numerical issues present during the computation of the bound based on the standard likelihood. Preliminary experiments on datasets with low contrast images, such as Frey Faces, show that the proposed bounds also allow the reconstruction of other types of images.

\section{Acknowledgements}
\label{Ack}
We would like to thank Andrei Ciuparu for useful discussions related to the training issues we encountered with the original ELBO and an anonymous reviewer for suggesting the reference \cite{Salimans2017}.

\nocite{langley00}

\bibliography{example_paper}
\bibliographystyle{icml2018}
\end{document}